\documentclass[sigconf]{acmart}
\AtBeginDocument{%
  \providecommand\BibTeX{{%
    \normalfont B\kern-0.5em{\scshape i\kern-0.25em b}\kern-0.8em\TeX}}}

\setcopyright{acmcopyright}
\copyrightyear{2018}
\acmYear{2018}
\acmDOI{XXXXXXX.XXXXXXX}

\acmConference[KDD '22]{Proceedings of the 28th
ACM SIGKDD Conference on Knowledge Discovery and Data Mining}{Aug 14--18,
  2022}{Washington, DC}
\acmPrice{15.00}
\acmISBN{978-1-4503-XXXX-X/18/06}

\graphicspath{ {./images/} }
\usepackage{subcaption}
\usepackage{hyperref}

\begin{document}

\title{MOSPAT: AutoML based Model Selection and Parameter Tuning for Time Series Anomaly Detection}


\author{Sourav Chatterjee}
\authornote{Corresponding Author}
\affiliation{%
 \institution{Meta}
 \city{Menlo Park}
 \country{USA}}
 \email{souravc83@fb.com}
 
 \author{Rohan Bopardikar}
\affiliation{%
 \institution{Meta}
 \city{Menlo Park}
 \country{USA}}
 \email{rohanfb@fb.com}
 
 \author{Marius Guerard}
\affiliation{%
 \institution{Meta}
 \city{Menlo Park}
 \country{USA}}
 \email{mariusguerard@fb.com}
 
 \author{Uttam Thakore}
\affiliation{%
 \institution{Meta}
 \city{Menlo Park}
 \country{USA}}
 \email{uthakore@fb.com}
 
 \author{Xiaodong Jiang}
\affiliation{%
 \institution{Meta}
 \city{Menlo Park}
 \country{USA}}
 \email{iamxiaodong@fb.com}

\renewcommand{\shortauthors}{Chatterjee et. al.}

\begin{abstract}
 Organizations leverage anomaly and changepoint detection algorithms to detect changes in user behavior or service availability and performance. Many off-the-shelf detection algorithms, though effective, cannot readily be used in large organizations where thousands of users monitor millions of use cases and metrics with varied time series characteristics and anomaly patterns. The selection of algorithm and parameters needs to be precise for each use case: manual tuning does not scale, and automated tuning requires ground truth, which is rarely available.

In this paper, we explore MOSPAT, an end-to-end automated machine learning based approach for model and parameter selection, combined with a generative model to produce labeled data. Our scalable end-to-end system allows individual users in large organizations to tailor time-series monitoring to their specific use case and data characteristics, without expert knowledge of anomaly detection algorithms or laborious manual labeling. Our extensive experiments on real and synthetic data demonstrate that this method consistently outperforms using any single algorithm.

\end{abstract}



\maketitle

\section{Introduction}
The problem of detecting anomalies and changepoints is central to a large number of engineering organizations in order to monitor system performance and 
reliability or understand sudden changes to user growth and retention.  In internet companies, anomaly and changepoint detection algorithms 
find particular applications in monitoring internet infrastructure, such as regressions in large codebases\cite{valdez2018real},
 monitoring network traffic\cite{kurt2018bayesian} and forecasting metrics\cite{taylor2018forecasting}. Such problems also occur in a variety of other 
 domains such as climate modeling\cite{manogaran2018spatial}, human activity recognition\cite{cleland2014evaluation},
  speech recognition\cite{panda2016automatic}, finance\cite{lavielle2007adaptive} etc.

There is a large body of work on anomaly\cite{braei2020anomaly} and changepoint detection\cite{aminikhanghahi2017survey}, that is often used in such scenarios. However, despite the large number of available algorithms, it is often challenging to solve anomaly detection problems in the real world. Time series generated from various sources are wildly different, and user intents for detection are often varied. No single detection method performs uniformly well for all scenarios, and the algorithm parameters are often nonintuitive and hard to tune manually without expert guidance. Finally, since ground truth is expensive to obtain, evaluating detection algorithms remains a challenging task\cite{van2020evaluation}. 

There has been little attention paid to the development of an AutoML approach to algorithm recommendation that works in the absence of ground truth data. In summary, users of time series anomaly detection algorithms are faced with the following problems:
\begin{itemize}
\item Time series data can vary wildly inside a large organization. Each team or application produces time series with differing characteristics. It is often unclear to the user which algorithms to pick for their purpose, and how to set thresholds.
\item Evaluating anomaly detection algorithms is hard due to lack of ground truth data. The process of manually labeling time series is labor intensive and impractical for individuals setting up algorithms. This makes a purely supervised algorithm selection approach impractical.
\item Users often do not have a deep expertise in anomaly detection. Individual algorithms are based on generative models and statistical assumptions, which are hard to know without explicit knowledge of those algorithms. Similarly, setting parameters often requires an expert knowledge of algorithms.  
\end{itemize}

 In a big organization, a large number of people are setting up monitoring for their metrics, and our aim is for users to set up monitoring without expert guidance (or without an analyst in the loop). This is a common problem in the time series domain. For example, the well known Prophet\cite{taylor2018forecasting} package attempts to solve this problem for forecasting at scale. With a similar goal in mind, for the anomaly detection space, in order to achieve the goal of automatic detector recommendations, we present an end-to-end framework, MOSPAT(\textbf{MO}del \textbf{S}election and \textbf{PA}rameter \textbf{T}uning) . Our main contributions are as follows:
\begin{itemize}
\item In the absence of ground truth, we propose a generative model to mimic the characteristics of the unlabeled data and inject different kinds of anomalies at known locations. This provides a training dataset for our supervised learning algorithm.
\item We implement a supervised learning approach, to learn the best detectors and parameters based on features of the time series.
\item We evaluate and benchmark a large number of existing time series detection methods, incorporated in our open source framework Kats\cite{kats2021}, against our AutoML approach and show significant improvements. We use both synthetic and real data for our evaluation.
\end{itemize}

A little more formally,  our real world use case consists of time series from K different sources. We assume that for each of the sources (e.g., a team in a large organization), a set of time series $T_k$ are generated from the same distribution $ T_k \sim D_k \forall k \in K $. Given a new time series from any of these sources, we would like to predict the algorithm and parameter combination, which could give the optimal performance for anomaly detection, without the need for manually labeled data from each of the K sources. In such problems, it is often useful\cite{choudhary2017runtime} to classify the types of anomalies we are interested in. In our case, we are interested specifically in three different kinds of anomalies, which we describe in detail in the paper. These are spikes, level shifts and trend shifts, shown pictorially in Figure \ref{figure:injection_illustration}. 

Our end-to-end algorithm consists of four important steps:
\begin{itemize}
\item \textbf{Offline Synthetic Data Preparation}: We generate a synthetic dataset with labeled known anomaly location, which mimic the actual data.
\item \textbf{Offline Training Data Preparation}: We use our synthetic data, to find the best algorithm for each dataset, and the best set of parameters, and use these as labels for our supervised learning task
\item \textbf{Offline Supervised Learning}: We learn supervised learning algorithms, which can predict the best algorithm and best parameters for a new time series.
\item \textbf{Online Inference}: This step takes place in real time, where, for a new time series, we predict the best algorithm, and then the best parameters for the algorithm.
\end{itemize}

Our approach is released in the open source library Kats\cite{kats2021}, and can be easily reproduced using publicly available data. 

\section {Related Work}
Evaluation is a hard problem in anomaly and changepoint detection. The literature often focuses on carefully designed synthetic datasets\cite{fearnhead2019changepoint}, and some standard datasets such as the Well Log data for level shifts\cite{knoblauch2018doubly, adams2007bayesian}. Some authors have created hand labeled datasets, for benchmarking algorithms. This includes the Numenta benchmark\cite{lavin2015evaluating} for anomaly detection and the Turing ChangePoint Dataset (TCPD)\cite{van2020evaluation}. However, each team produces time series data that is very specific to their own needs, and thus using gold standard benchmark datasets would often lead to suboptimal results for a team, looking to set up anomaly detection in the real world.

The Numenta benchmark evaluation\cite{lavin2015evaluating}  designs a custom scoring strategy by assigning different scores to true positives and false positives and penalizing late detection. TCPD\cite{van2020evaluation} uses clustering and classification based metrics. Our evaluation metrics correspond most closely to the classification based approach described here.

Previous surveys of anomaly detection algorithms have tried to classify anomalies into different subdomains. For example, Choudhary et. al.\cite{choudhary2017runtime} classifies anomalies as point anomalies, change detection (level and variance changes), pattern anomalies and concept drift. Cook et. al.\cite{cook2019anomaly} makes a similar classification of point anomalies, contextual anomalies and pattern Anomalies. 

A number of time series anomaly detection algorithms are used in the industry, such as a framework from Microsoft\cite{ren2019time}, Merlion from Salesforce\cite{bhatnagar2021merlion}, Luminaire from Zillow\cite{chakraborty2020building}. A number of similar issues such as the lack of labels and generalization\cite{ren2019time}, the need for AutoML\cite{bhatnagar2021merlion} and synthetic anomalies\cite{chakraborty2020building} are pointed out in these works coming out of the industry. 

AutoML has been used widely in machine learning as an approach to improve performance metrics\cite{ledell2020h2o}. In the context of time series, AutoML methods have found large acceptance in forecasting\cite{zhang2021self} and time series data cleaning\cite{shende2021cleants}. Some existing end-to-end anomaly detection frameworks, such as Merlion\cite{bhatnagar2021merlion}, PyODDS\cite{li2020pyodds}, TODS\cite{lai2020tods} have employed AutoML methods. Beyond using AutoML to select the best parameters, time series anomaly detection algorithms have learned ensembles\cite{jamil2021ensemble} and used human inputs to select best detection algorithms\cite{freeman2019human}. However, these algorithms assume the existence of labeled data to learn the best parameters and do not build an end-to-end system which can work in the absence of labeled data. 

Finally, synthetic time series generation has become important in recent years due to an emphasis on privacy preserving approaches. TimeGAN\cite{yoon2019time} is a deep learning based approach to generate synthetic datasets. However, such synthetic data generation methods cannot produce time series with labeled anomalies of various kinds.

\section{Detection Algorithms}
We consider a number of individual detection algorithms we choose from. These algorithms are a part of the open source package Kats\cite{kats2021}, which the authors are involved in.  We use well known algorithms which are sometimes modified to capture the peculiarities of time series encountered in our applications. For example, in internal companies, business time series often show strong seasonality, holiday effects etc.\cite{taylor2018forecasting}.   

Below, we give a brief description of the algorithms. Along with their description, we also mention the hyperparameters for each algorithm. This is important, since our end to end system suggests the best hyperparameters for each algorithm.

\begin{itemize}
\item \textbf{Outlier}:  This is a faithful implementation of the \textit{outlierts()} function in the forecast package in R\cite{hyndman2008automatic}. First, we use STL decomposition, to decompose the time series into a trend and seasonality and a residual. We de-trend and de-seasonalize the time series. The residual is scaled by the IQR and an user specified multiplier, and reported as the output score.The hyperparameter for this algorithm is the IQR multiplier, i.e. how many multiples of interquartile range should be considered an outlier. Intuittively, we think this method will be useful for catching spike anomalies.
\item \textbf{CUSUM}: This is a slight modification of the CUSUM algorithm\cite{page1954continuous}. CUSUM refers to Cumulative sum, and refers to calculating the cumulative sum of the normalized observations $z_t$, where
\begin{equation}
z_t = \frac{x_t - \bar{x}}{\sigma_x}
\end{equation}
At each time step, we calculate a cumulative sum ($\omega$ is the likelihood function)
\begin{equation}
	S_{t+1} = max(0, S_t + z_n - \omega)
\end{equation} 

In our settings, we often look at time series, that show a deviation from a normal behavior and return back to normal. CUSUM is specially designed to catch such cases. Since seasonality is important in our applications, we also remove seasonality by using STL decomposition before applying CUSUM. The hyperparameters for this algorithm consist of the historical data windows and a boolean flag indicating whether seasonality should be removed.

\item \textbf{Statsig}: This is a detector, that uses a t-test to determine if changes between a control and test window are significant. This is a detector preferred by many users because of its simplicity and interpretability. The hyperparameters for the algorithm consist again of the history and test windows. 
\item \textbf{BOCPD}(Bayesian Online Change-Point Detection): This is a faithful reproduction of the Bayesian Online Changepoint Detection algorithm \cite{adams2007bayesian}. BOCPD adopts a Bayesian approach to changepoint detection. It assumes a prior probability for changepoint at time t, as $p(r_t)$. and after observing data $x_{1:t}$, it calculates the posterior probability $P(r_t |  x_{1:t})$. BOCPD basically assumes an underlying predictive model (UPM), $P(x_t | x_{1:t-1})$ corresponding to the generative model of the data, and uses a recursive equation to calculate $P(r_t |  x_{1:t})$. based on the UPM term $P(r_{t-1} |  x_{1:t-1})$.  The hyperparameter consists of the prior probability of a changepoint $p(r_t)$. BOCPD is shown to be one of the best performing changepoint detection algorithms, in a comprehensive evaluation of changepoint detectors\cite{van2020evaluation}.

\item \textbf{MKDetector}: This algorithm detects trend changes, based on the non-parametric Mann-Kendall test\cite{mann1945nonparametric, kendall1948rank}. Its null hypothesis is that there is no monotonic trend in the time series, and its alternate hypothesis the opposite. A sudden change in the value of the test statistic indicates a changepoint.
The Mann-Kendall test is expensive, and is performed within a  window of data from the current point. The hyperparameters are the data window and an indicator on whether seasonality should be removed. 
\item \textbf{Prophet}: Prophet\cite{taylor2018forecasting} is widely used in the industry as a forecasting algorithm. Prophet is designed keeping in mind the characteristics of business time series which have strong seasonality, holiday effects, etc. Although Prophet is a forecasting algorithm, it detects changepoints during the course of forecasting. These detected changepoints have been used in a number of changepoint detection applications\cite{tran2020interpreting, van2020evaluation}. We use Prophet for trend changepoint detection.

\end{itemize}

As we have mentioned before, our specific interest is in three different kinds of anomalies, spikes, level shifts and trend shifts. We have included these particular algorithms, to capture these three kinds of anomalies. Intuitively, and based on the statistical models used, we believe that the following algorithms should best capture each type of anomaly:
\begin{itemize}
\item Spike: Outlier
\item Level Shift: CUSUM, BOCPD, StatSig
\item Trend Shift: Prophet, MKDetector
\end{itemize}

Each algorithm returns an anomaly score, of the same length as the time series. Anomaly scores can take different values, depending on the algorithm. In general, scores of larger absolute value indicate higher likelihood of an anomaly, or an anomaly of a larger magnitude. For each algorithm, in addition to its specific hyperparameters, we also treat the high and low (upward and downward) thresholds of the scores as hyperparameters, and search over this space.

\section{Evaluating Detection Algorithms}
Our evaluation method closely follows that of Burg et. al.\cite{van2020evaluation}, who take this view of changepoint detection algorithms as a classification problem. 
Our goal will be to define precision P and recall R, and finally combine them into one single score,  which corresponds to the $F_\beta$ measure.
\begin{equation}
F_\beta = \frac{(1 + \beta^2) PR}{\beta^2P + R}
\end{equation}
We use $\beta=1$, which corresponds to the F1-score.

The definition of precision and recall are modified in order to accomodate for a margin of error around the true change point\cite{killick2012optimal,van2020evaluation}. This allows for minor discrepancies and a small detection delay. However, similar to Killing et. al.\cite{killick2012optimal}, we make sure that if multiple changepoints are detected inside the margin, we only count one as a true positive.
Let $\chi$ be the set of all changepoint locations detected by the detection algorithm. and $\Gamma$ be the set of ground truth changepoint locations. We define true positives $TP(\Gamma, \chi)$ as  those $ \tau \in \Gamma$ for which $\exists x \in \chi$ such that $|\tau - x| \leq  M$. Only a single $x \in \chi$ can be matched to a single $\tau \in \Gamma$.

\begin{equation}
P = \frac{|TP(\Gamma,\chi)|}{|\chi|}
\end{equation}

and 
\begin{equation}
R = \frac{|TP(\Gamma,\chi)|}{|\Gamma|}
\end{equation}

As a convention, we include t=0 as a trivial changepoint, which avoids the definition being undefined in cases where there are no changepoints.

In addition to the above, of particular interest to us is the number of false positives. We define as the set of all positives which are not in the the margin of a true positive
\begin{equation}
FP(\Gamma, \chi) = \{x \in \chi \ni |x-\tau|>M  \forall \tau \in \Gamma\}
\end{equation}

\section{Generating Labeled Data}

\begin{figure}
\includegraphics[width=\linewidth]{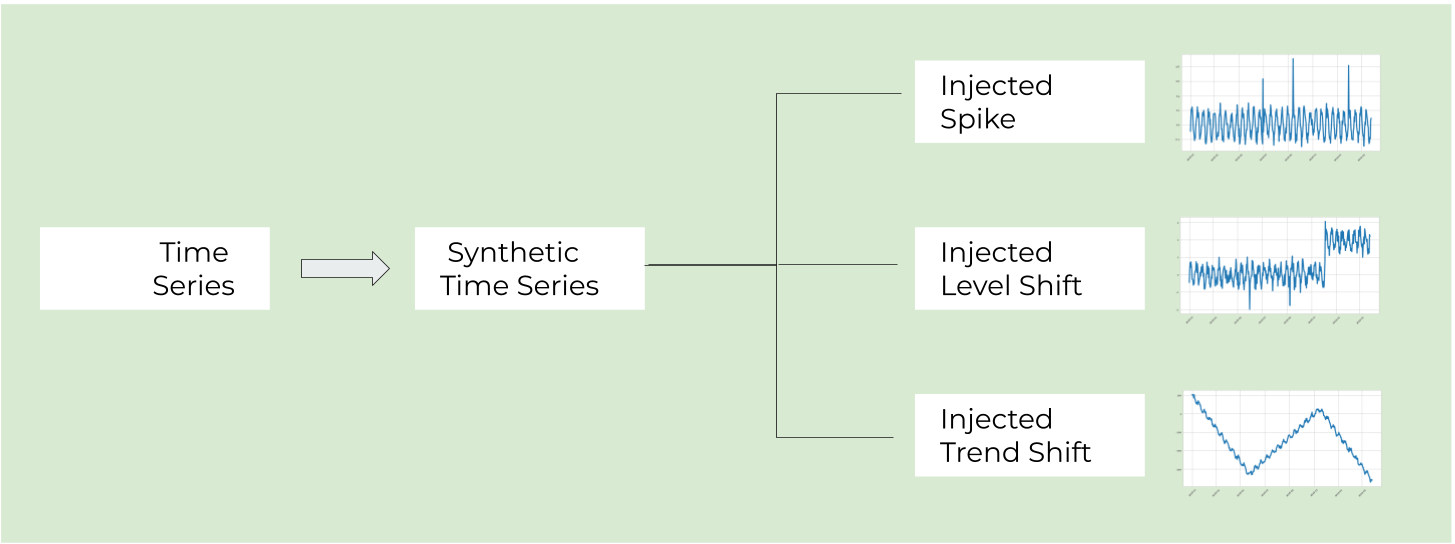}
\caption{Workflow of generating training data, based on user provided unlabeled data}
\label{figure:injection_illustration}
\end{figure}
A huge handicap to creating and evaluating anomaly detection models is the absence of ground truth data. Typically, the user input is a set of time series, which may or may not contain anomalies. These time series are unlabeled, i.e., the location and the presence of anomalies are unknown. We are interested in the problem of detecting anomalies from other time series generated from the same distribution, i.e., having similar characteristics. We produce labeled training data from this unlabeled time series data in two steps. In the first step, we construct a simulated time series that has similar characteristics to the real time series but does not contain anomalies. In the second step, we inject appropriate anomalies on this simulated time series. The workflow is illustrated in Figure \ref{figure:injection_illustration}. 

\subsection{A Simulated Time Series}
First, we produce a simulated time series from the unlabeled time series data, which may or may not contain anomalies. To do this, we decompose the original time series with STL decomposition\cite{cleveland1990stl}. STL decomposition divides the time series into three parts: the trend, the seasonality, and the residuals.  
We assume that the anomalies in the real time series will appear in the trend component, or as outliers in the residual component. We construct a synthetic time series additively by adding up random noise, with the same mean and standard deviation as the residual of STL. Then, we add the seasonality component. The simulated time series is supposed to reflect the characteristics of the real time series, but not the anomalies. This is useful because:
\begin{itemize}
\item We can generate an unlimited amount of synthetic data, often larger than the size of the original dataset
\item We can eliminate the unknown anomalies in our original data. Since we do not know the location of these anomalies, this provides a problem when planning to use this data as training data.
\end{itemize}

\subsection{Injected Anomalies}
In order to create anomalies on the synthetic data, first, we choose the locations of anomalies.  The location of anomalies are assumed to be independent. Thus, the intervals between anomalies are random draws from a geometric distribution as follows:
\begin{equation}
\tau_s - \tau_{s-1} = G(\frac{1}{\tau_{dist}})
\end{equation}

where $\tau_{dist}$ is the average interval length. 

In the case of level and trend shifts, we assume that the time series is divided into segments $\tau_1 < \tau_2 < \tau_3 <...\tau_k$ where $\tau_s$ is drawn as above. 

Once we have the locations of the anomalies, we then inject three different kinds of anomalies: spikes, level shifts, and trend shifts. Each time series is injected with one kind of anomaly.
\begin{itemize}
\item \textbf{Spike}: The z-score of spikes is first drawn from a normal distribution. The direction of spikes is then drawn from a Bernoulli distribution.
\begin{equation}
	S_{spike} \sim Ber(p_{spike})
\end{equation}

\begin{equation}
	Z_{spike} \sim N(\mu_{spike}, \sigma_{spike}^2) 
\end{equation}

Finally, we find the empirical standard deviation of the time series data $\hat{\sigma}$ and inject spikes in locations indicated by $\tau_s$
\begin{equation}
	x_{\tau_s} = x_{\tau_s} + (-1)^{S_{spike}[s]} \left | Z_{spike}[s] \right | \hat{\sigma}
\end{equation}

\item \textbf{Level Shifts}:  As indicated above, in this case, $\tau_s$ indicates a changepoint. and each segment is defined by a level $\mu_s$. Therefore, the points in the time series are generated as
\begin{equation}
x_t \sim N(\mu_s, \sigma_L) \forall t \in (\tau_{s-1}, \tau_s)
\end{equation}

The magnitude of the level shift between two consecutive changepoints is drawn from a normal distribution.

 We look at two different commonly observed level shifts. In a number of applications the time series shows alternate level shifts. As an example, efficiency regressions in infrastructure systems\cite{valdez2018real}  are  characterized by instances where the CPU utilization or query per second (QPS) patterns of a function or endpoint experience an unexpected increase over its prior baseline. At the end of the regression, the function comes back to its baseline value. We simulate this by injecting alternate level shifts. We also simulate a more general level case of a level shift, where the sign of the shift is a draw from a Bernoulli distribution as before.
 
 In the first case, we can write:
 \begin{equation}
\mu_s \sim (-1)^s \left |N(\mu_{s-1}, \sigma_\mu) \right | 
\end{equation}

In the second case,

\begin{equation}
	S_{level} \sim Ber(p_{level})
\end{equation}

 \begin{equation}
\mu_s \sim (-1)^{S_{level}[s]} \left |N(\mu_{s-1}, \sigma_\mu) \right | 
\end{equation}

\item \textbf{Trend Shifts}:  As above,$\tau_s$ indicates a changepoint. and each segment is defined by a trend $\beta_s$.
Therefore, inside a segment, the points in a time series are generated as:
 \begin{equation}
 x_t  \sim N(\beta_s t, \sigma_T)
 \end{equation}
 This is equivalent to representing each segment as a straight line with added Gaussian noise. As before, we consider alternate shifts in this case. This is the most commonly observed scenario, where a metric is increasing and suddenly starts decreasing due to some abnormality, which is an event of interest to be detected. In this case, 
 
\begin{equation}
\beta_s \sim (-1)^s \left |N(\beta_{s-1}, \sigma_\beta) \right | 
\end{equation}
 
\end{itemize}

\begin{figure}
\includegraphics[width=\linewidth]{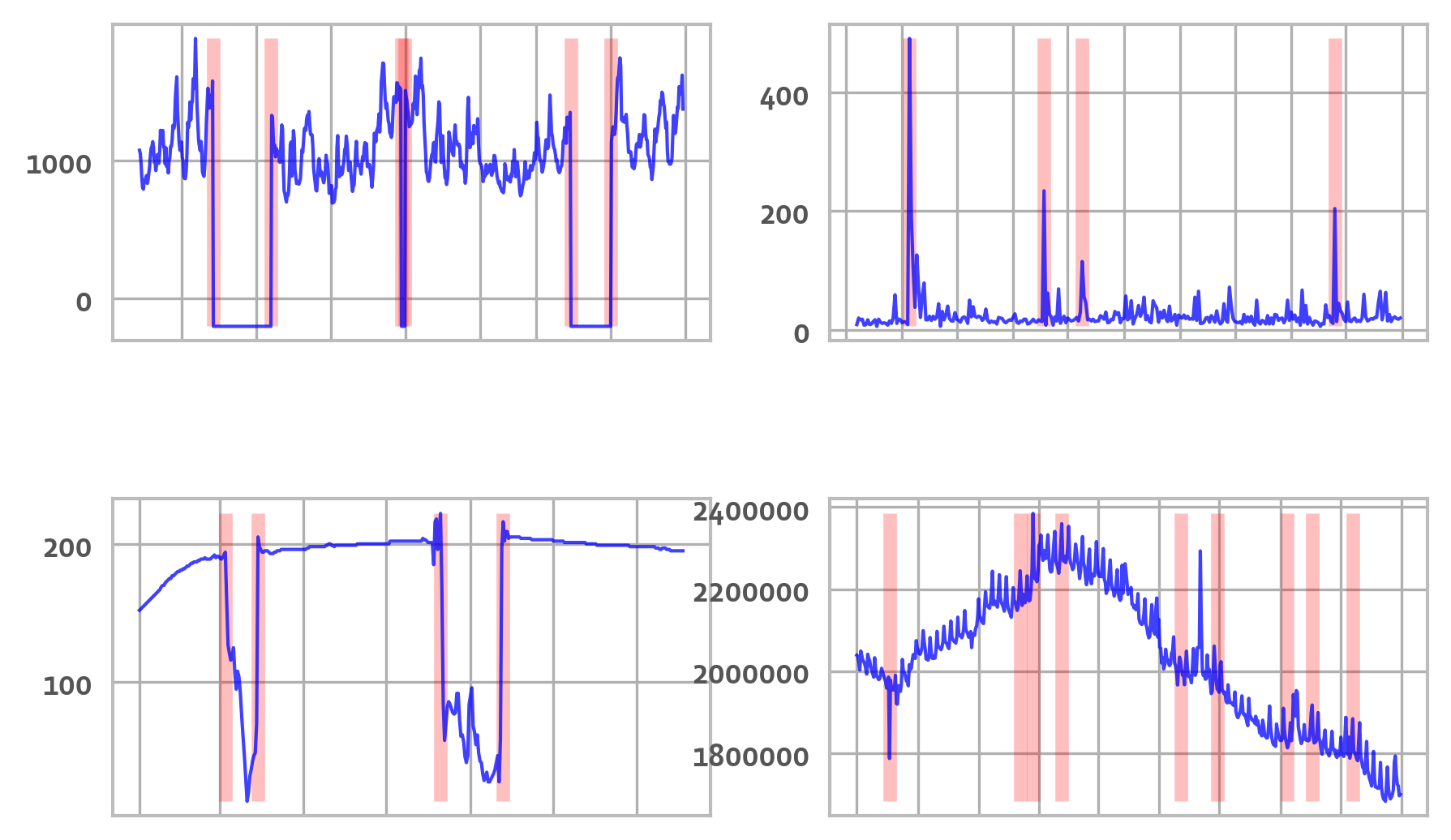}
\caption{Original time series produced by injecting anomalies to real time series data}
\label{figure:real_example}
\end{figure}

\begin{figure}
\includegraphics[width=\linewidth]{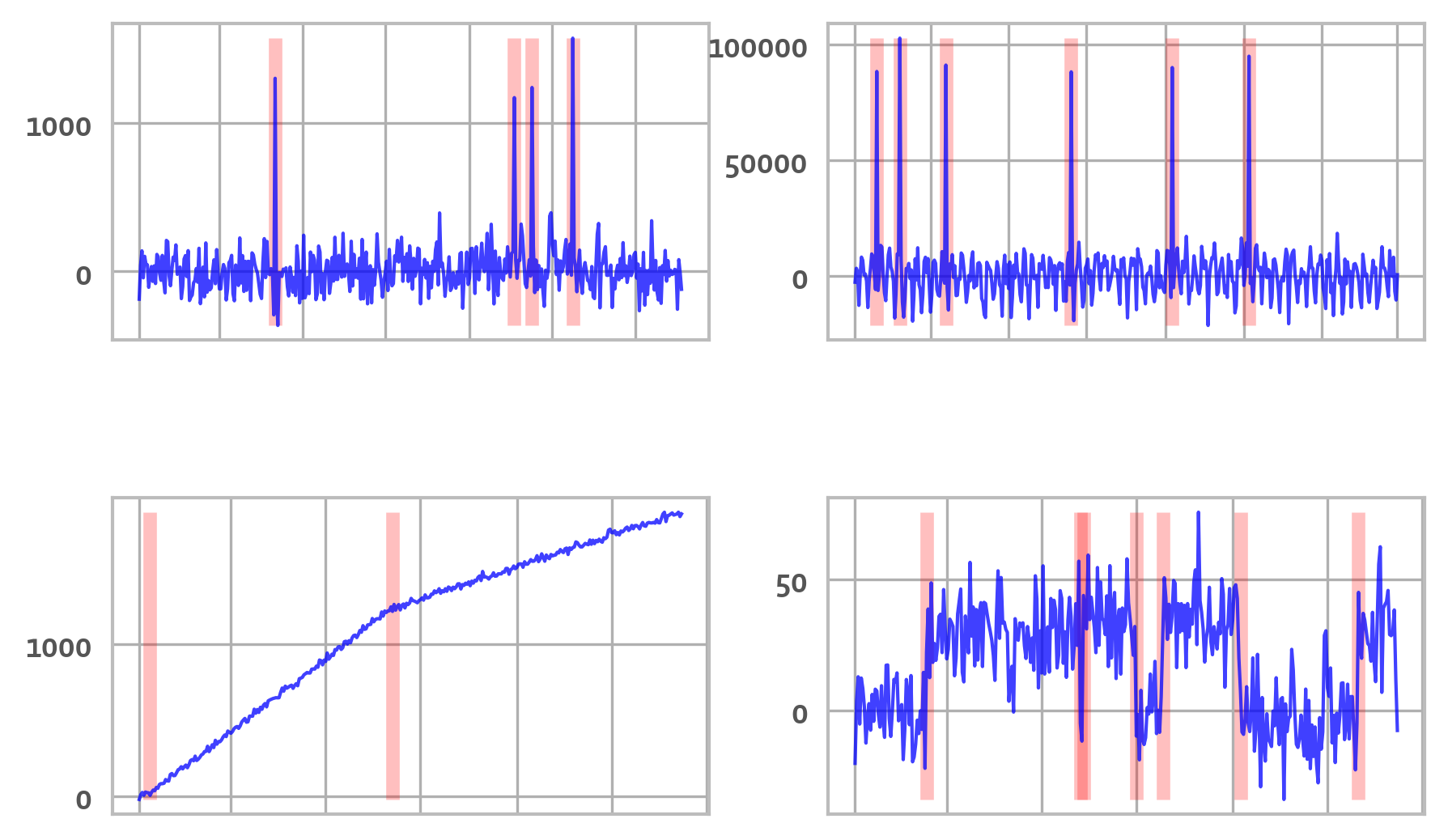}
\caption{Simulated time series produced from the original time series data}
\label{figure:simulated_example}
\end{figure}

The parameters for these distributions are chosen, by making reasonable estimates from empirical observations of time series data in our domain.
Figure \ref{figure:real_example} shows four randomly chosen examples of real time series data used in our experiments (the data is described in detail in Section 7.3). Figure \ref{figure:simulated_example} shows four randomly chosen examples of labeled training data produced by our generative model. As it can be seen, the generative model captures the salient features of the time series, such as spikes, level/trend shifts and seasonality. To further investigate this, we performed dimensionality reduction using T-SNE, which we describe in detail in Section 7.

\section{Automated Machine Learning}

\begin{figure}
\includegraphics[width=\linewidth]{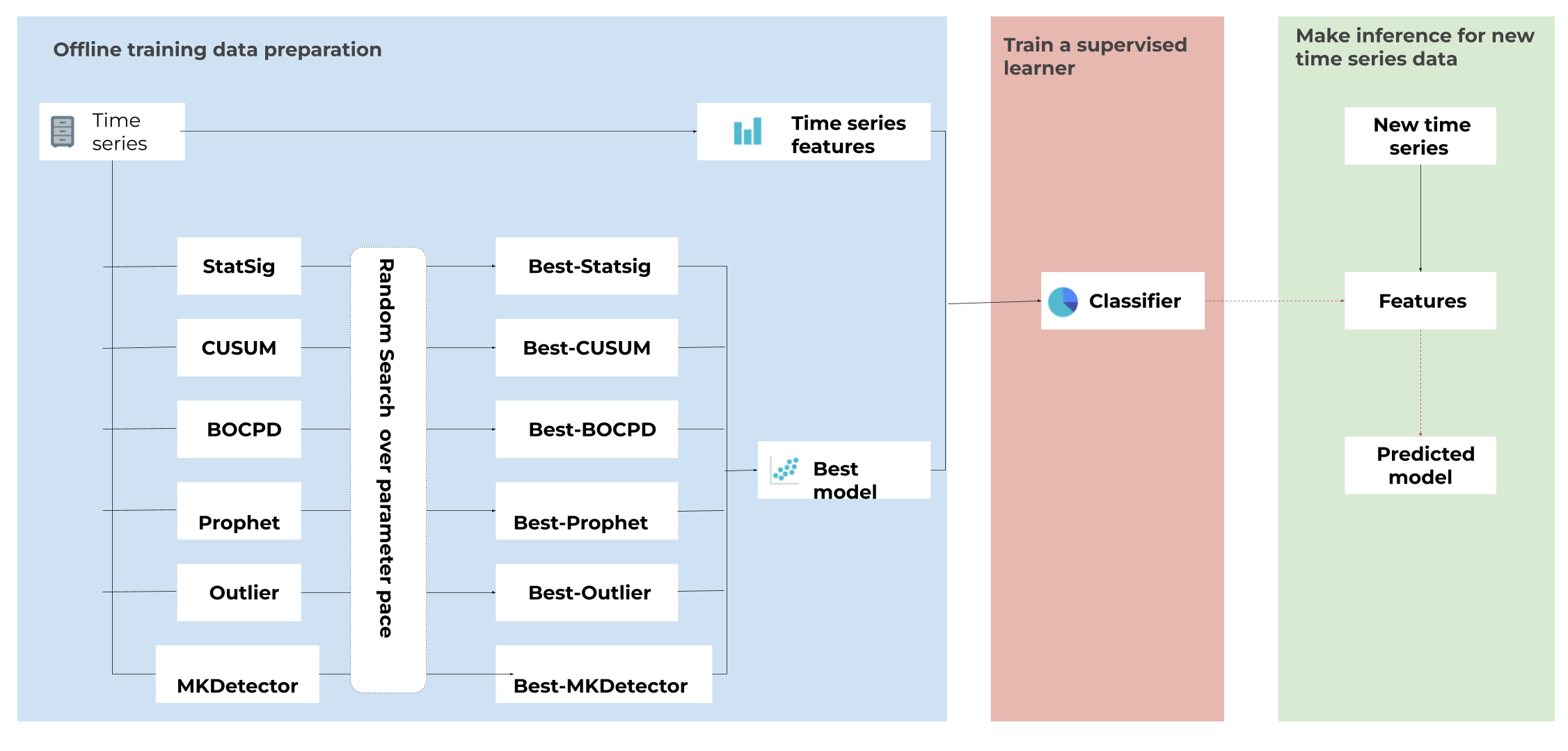}
\caption{Workflow of training AutoML algorithm}
\label{figure:automl}
\end{figure}

In the next step of our end-to-end system, we use the labeled anomaly data we generated, to learn the best algorithm and parameters for a given time series. This AutoML approach is undertaken in two stages. Our first model attempts to do model selection, and then a second model is used to find the best hyperparameters. The workflow is illustrated in Figure \ref{figure:automl}.  Our approach to AutoML is similar in nature to that described for time series forecasting\cite{zhang2021self}, with notable differences on evaluation procedure and algorithms used. Our steps to prepare the training data and algorithm training, although time consuming, are done offline and are a one time procedure for a new team setting up their anomaly detection. Inference and end to end anomaly detection is done online and does not take substantial wall time, so it can be done in real time by users. 
Here are the steps we take for automated machine learning:
\begin{itemize}
\item \textbf{Building the Dataset: } First, we build a training dataset for this model. For each simulated time series with an injected anomaly, we try all the base models; for each model, we perform a random search over its parameter space to find the set which optimizes the chosen metric. For our cases, we choose the F-score as the metric, but our framework supports a number of different metrics, such as precision, recall and detection delay. Preparation of the training set is the most time consuming portion of the problem. We store the best model and the best hyperparameters for each model. The best model serves as the label for our supervised training procedure. When multiple models give the same F-score for a given time series we randomly choose one, as the best model. 
\item \textbf{Feature Extraction: } We extract features for each time series, which are used as features for our supervised learning algorithm. These features include the characteristics of the time series such as trend, seasonality, autocorrelation, nonlinearity, etc. We used 40 features in our final training algorithm. 
\item \textbf{Supervised model for Model Selection: } We use a supervised learning algorithm, in this case a random forest, to predict the label for a given time series. Thus, for a given time series, we can predict the best performing anomaly detection model. The choice of classification algorithm is flexible, in our framework. The user can decide to choose other algorithms, such as logistic regression, GBDT or k-nearest neighbors. 
\item \textbf{Hyperparameter Tuning: } We take a machine learning based approach to hyperparameter tuning. Traditional hyperparameter tuning approaches, such as grid search and random search are too expensive for our case, which attempts to provide near real time information to users. We train a multi-task neural network, similar to previous work in forecasting\cite{zhang2021self}, which takes the time series features as inputs and predicts the best hyperparameters for a given algorithm. The multi-task neural network consists of shared layers and task specific layers, one for each hyperparameter. The losses are the average of the loss for the neural network for each hyperparameter. The reason we require a multi-task neural network is because we often have a mixture of categorical and numerical hyperparameters (for example, in CUSUM the scan window and historical window are numerical parameters, but the flag on whether to remove seasonality is categorical) and require some of the networks to minimize a classification loss and some to minimize a regression loss. Details on the architecture of the network can be found in Zhang et. al.\cite{zhang2021self}. For each algorithm, we train a separate neural network. At inference time, once an algorithm is chosen, we predict the best set of hyperparameters by using the multi-task neural network corresponding to the algorithm. 

\end{itemize}

\section{Results and Discussions}
In this section, we present our experimental results with synthetic and real datasets. Our major challenge was the lack of labeled datasets, which correspond to our use case. Our real world use case consists of time series from multiple sources, with the time series from each source generated from the same distribution $ T_k \sim D_k$. Given a new time series from one of the sources, we would like to predict the algorithm and parameter combination, that would give the optimal performance. Therefore, in our experiments, we combine various data sources, simulate different kinds of anomalies, and compare our AutoML approach to using a single base algorithm consistently for all time series. In order to test the efficacy of the hyperparameter tuning, we compare the results of choosing the hyperparameter using the multi-task neural network versus choosing a random hyperparameter from the hyperparameter space of an algorithm, which we define. This setup  most closely resembles how users would invoke our system. Given our challenges with labeled data, we perform experiments in three different setups that correspond to increasing levels of realism in the data:
\begin{itemize}
\item Synthetic time series data, overlaid with anomalies
\item Real time series data, overlaid with anomalies
\item Real, labeled time series data
\end{itemize} 

In the following sections, we will make a distinction between the input dataset, which is a user input of unlabeled data (they are labeled in the experiments for evaluation purpose), and training data, which is labeled data, composed by our end-to-end system described in Section 5.  

\begin{figure}
	\begin{subfigure}{\linewidth}
	\includegraphics[width=0.9\linewidth]{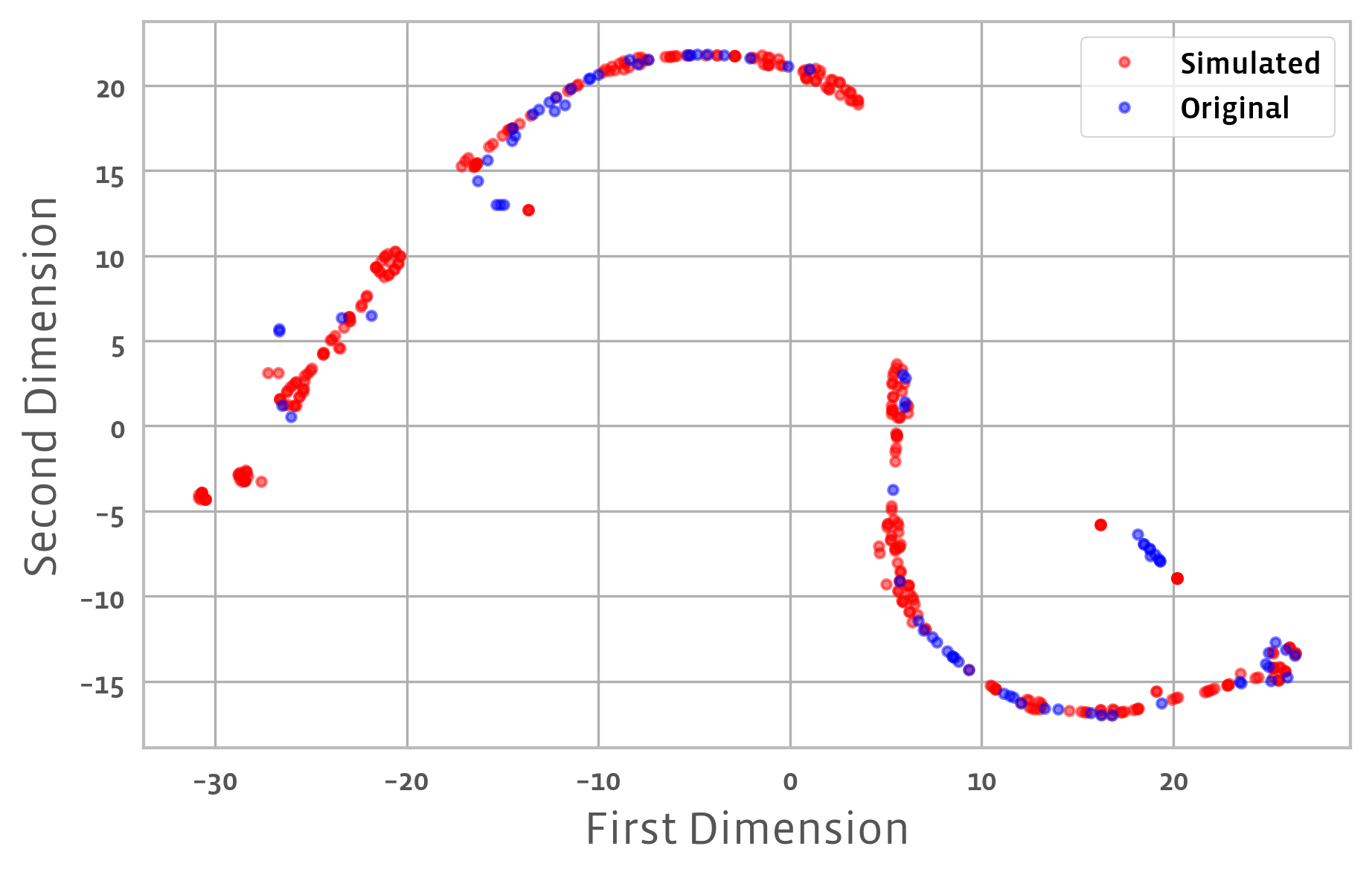}
	\caption{}
	\end{subfigure}
	
	\begin{subfigure}{\linewidth}
	\includegraphics[width=0.9\linewidth]{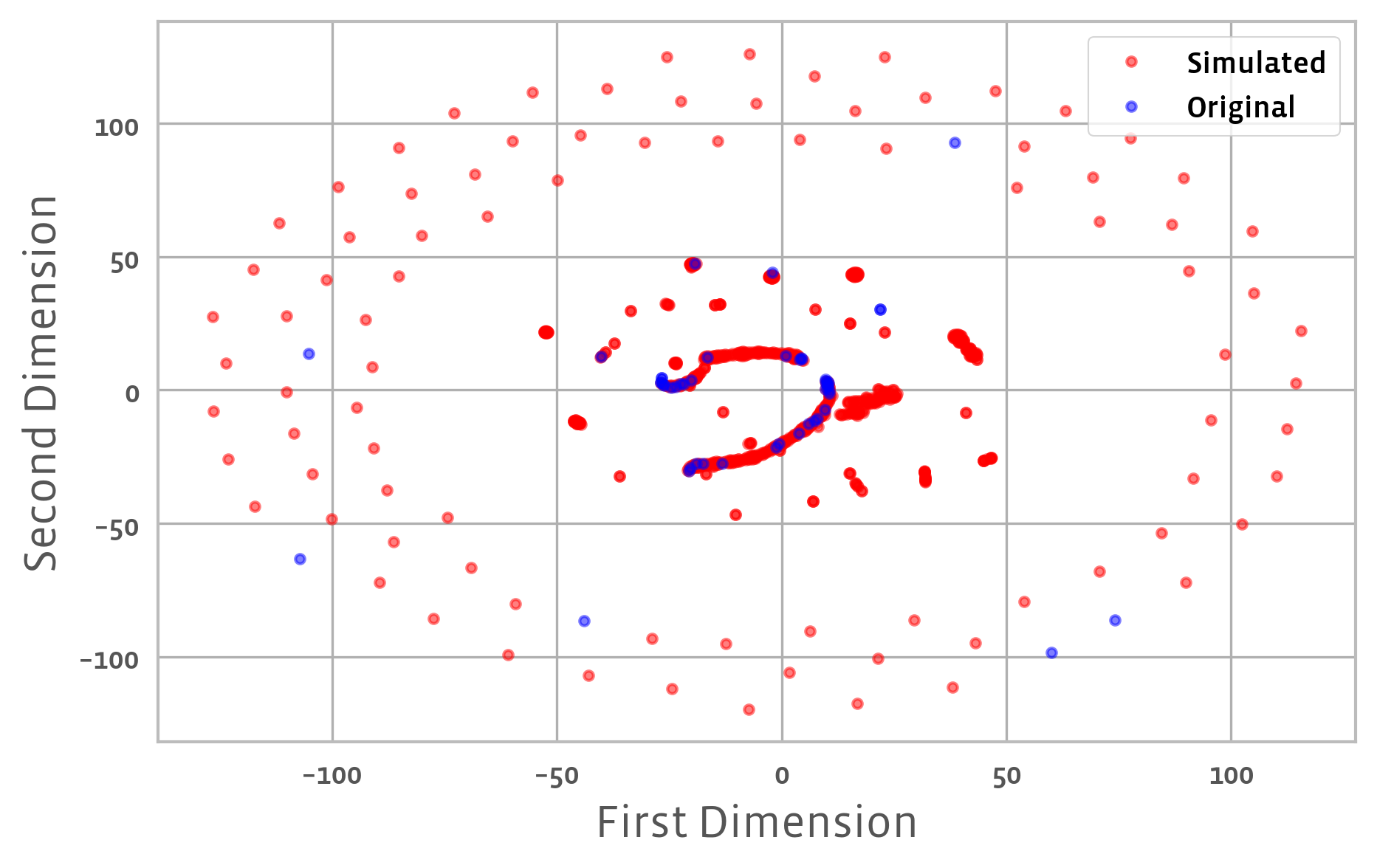}
	\caption{}
	\end{subfigure}
\caption{T-SNE plot, which shows that the training data shares the same distribution as the original input data for (a) injected anomalies overlaid on real data and (b)real data}
\label{figure:tsne}
\end{figure}

\subsection{Synthetic Data}
Our basic approach is to start with a small set of time series without anomalies, which is generated according to a given generative model. We then inject various kinds of anomalies on this data to produce a larger set of data, for which we know the locations of the anomalies. The system is then fed the unlabeled time series data, and the known labels are used only for evaluation. We start with a set of 18 time series. Ten of them are produced by adding trend noise and weekly seasonality additively, five of them are produced from an ARIMA model, two of them from i.i.d normal, and one from an i.i.d. t-distribution (degree of freedom 5). For the ARIMA model, we use $ARIMA(p=2, d=\alpha, q=2)$, where $\alpha \sim Ber(0.5)$.  The length of each time series is a draw from a Poisson distribution with a mean parameter 450. 

On this set of 18 time series, we inject anomalies using our simulator to produce a larger dataset of 432 time series, which is our input dataset. We use 80 percent of the data for training and set aside the rest 20 percent purely for evaluation of the algorithms.  Using the training set, we create labeled synthetic data and use this synthetic data to train a supervised learning algorithm. 
Unlike classification problems, we are not directly interested in the score of the classification model itself. Instead, we are interested in evaluating the end to end workflow. We evaluate the workflow in two steps, for each time series in the evaluation set, 
\begin{itemize}
\item We predict the best model using the trained supervised model.
\item We use either random hyperparameters or our multi-task neural network to predict the hyperparameters.
\end{itemize}  
We then calculate the F-score for our model and the chosen set of hyperparameters.

The average F-scores on the evaluation data are reported in Table \ref{table:synthetic}.  We show optimized F-scores, which are obtained when the parameters of the algorithm are chosen using AutoML, along with random F-scores, where the parameters are a random choice from the parameter space. We observe that AutoML performs better, in terms of F-score than using any particular algorithm consistently, for all the data. AutoML results are also considerably improved by hyperparameter tuning. Hyperparameter tuning seems to improve some algorithms such as BOCPD a lot (more than 50 percent), whereas MKDetector  or Prophet seems to not be not helped by hyperparameter tuning. This is also observed in the experiments with real and injected time series data in the next sections. Given that AutoML is consistently helped by hyperparameter tuning, we assume that the cases where hyperparameter tuning does not work, are the ones which are not appropriate for this detector type, for example spikes in the case mkdetector(which is deployed to capture trend shifts). Thus, our hyperparameter tuning, although does not work for these cases, it works well for AutoML, which is our main objective.  

\begin{table}
\begin{center}
\begin{tabular}{|c|c|c|}
\toprule
Method & F-Score(Optimized) & F-Score(Random)\\
\midrule
AutoML & \textbf{0.754} & 0.573\\
Outlier & 0.628 & 0.554\\
CUSUM & 0.477 & 0.455\\
BOCPD & 0.577 & 0.223\\
Statsig & 0.372 & 0.202\\
MKDetector & 0.374 & 0.430\\
Prophet & 0.4 & 0.418\\
\bottomrule
\end{tabular}
\end{center}
\caption{Comparison of AutoML with baseline algorithms on Synthetic data}
\label{table:synthetic}
\end{table}

\subsection{Real Data with Injected Anomalies}
Our real datasets consist of 18 publicly available time series datasets, 13 from the Google Mobility dataset\cite{google2021mobility} and 5 from the Bee-Dance dataset\cite{oh2008learning}. Google mobility data summarizes time spent by their users each day after Feb 6, 2020 in various types of places, such as residences, workplaces, and grocery stores. The values in the time series represent percentage change in mobility compared to the baseline measured on Feb 16th, 2020.   This data contains strong weekly seasonality, a characteristic shared in a lot of applications we care about. The bee-dance dataset\cite{oh2008learning} records the pixel locations in x and y dimensions and angle differences of bee movements. Ethologists are interested in the three-stages bee waggle dance and aim at identifying the change point from one stage to another. By contrast to the google mobility data, it does not contain any seasonality effects. 

Using real datasets instead of synthetic datasets allows us to simulate a more realistic setting, while still maintaining a large enough dataset. We focus on segments of time series where there are no huge spikes. In case of anomalies in the time series, we label the anomalies. Based on the 18 time series data, we create 432 time series by injecting anomalies according to the procedures described in Section 3. We consider the concatenation of the anomalies in the original time series as well as injected anomalies as the real labels. We then proceed to evaluate in the same way as the purely synthetic data. 

First, we compare the distributions of the generated training data to the evaluation data. We perform dimensionality reduction of the time series using T-SNE and project it to two dimensions in Figure \ref{figure:tsne}(a). We observe a high level of overlap between the two distributions, which indicates that the generated time series data is able to mimic the characteristic of the input data. This confirms the validity of our generative model in this simple case.

Table \ref{table:injected} shows the results of the average F-score, both for a choice of random hyperparameters and the optimal hyperparameters. We are able to get an average F-score of 0.720 using AutoML,which shows an improvement of 12 percent over the best performing single algorithm, BOCPD. Hyperparameter tuning is shown to improve the performance of AutoML in this case. 

Further, we want to analyze whether these selected models correspond to what we would expect intuitively. As mentioned before, the algorithms were developed in order to tackle specific kinds of anomalies. Outlier is supposed to be good at catching spike anomalies, Statsig/BOCPD/CUSUM good at catching level shifts, and MKDetector/Prophet are designed to catch trend shift anomalies. Since in the purely synthetic and real-injected data, we have manually injected various kinds of anomalies, we can check whether the best predicted algorithm (for the evaluation dataset) corresponds to our intuition. Tables \ref{table:intuitive_synthetic} and \ref{table:intuitive_injected} show the results for synthetic and injected data respectively. We show the number of examples of each injection type for which an algorithm was chosen to be the best by the AutoML procedure. Largely, our intuitions are true, and we can see Outlier is good at identifying spikes, BOCPD at level shifts, and Prophet and MKDetector are good at trend changes. However, these assumptions are not always true. For instance, in Table \ref{table:intuitive_synthetic}, we observe that Outlier is also quite good at capturing trend changes. Thus, AutoML, while intuitively directionally correct, does not depend on such heuristics and is able to select algorithms and improve upon the intuition. 

\begin{table}
\begin{center}
\begin{tabular}{|c|c|c|}
\toprule
Method & F-Score(Random) & F-Score(Optimized)\\
\midrule
AutoML & \textbf{0.720} & 0.682\\
Outlier & 0.491 & 0.550\\
CUSUM & 0.563 & 0.576\\
BOCPD & 0.638 & 0.287\\
Statsig & 0.486 & 0.321\\
MKDetector & 0.483 & 0.530\\
Prophet & 0.497 & 0.5\\
\bottomrule
\end{tabular}
\end{center}
\caption{Comparison of AutoML with baseline algorithms on injected anomalies on Real Data}
\label{table:injected}
\end{table}

\subsection{Real Data}
While our previous experiments on synthetic data allow us to get large sample sizes, they may not be completely realistic. Our end-to-end approach also gets the advantage that the anomaly injection process used to construct the input dataset is also similar to that used to construct the labeled training data. Hence, we test our algorithm on purely labeled real data. This is expensive to obtain, but is the most golden test for our AutoML procedure.

Our real datasets consist of 106 time series from a mix of 4 different sources, which best replicate our user environment. These data sources are also publicly available, making it easy to replicate our results. Although the numbers seem small, compared to examples used in supervised learning, anomaly detection data is much more expensive to collect and label. For comparison, the previous most comprehensive study of changepoint detection algorithm\cite{van2020evaluation} had used 37 labeled time series. We have collected the time series data, and hand labeled the anomaly locations. The data sources are
\begin{itemize}
\item \textbf{Scanline}: This is a univariate time series of pixel values, which is a horizontal scan line of
image no. 42049 from the Berkeley Segmentation Dataset\cite{martin2001database}. It has been previously used for changepoint evaluation\cite{van2020evaluation} and shows characteristic level shifts between image segments. We scan horizontally at multiple locations to obtain 20 time series.
\item \textbf{Air Quality}: This consists of time series from the air quality dataset\cite{de2008field} in the UCI Machine Learning repository\cite{Dua:2019}. The data is collected hourly and has a daily seasonality, characteristic of a number of time series that are modeled in internet companies. This dataset has been used previously for forecasting\cite{iskandaryan2020air}. 
\item \textbf{Web Traffic}: This consists of web traffic to individual wikipedia pages\cite{kaggle2017webtraffic}, released by Google. This dataset has been used extensively for time series forecasting benchmarking\cite{petluri2018web}.
\item \textbf{HAR}: The HAR dataset \cite{anguita2013public} in the UCI Machine Learning repository\cite{Dua:2019} provides human activity information collected by portable accelerometers, when participants different activities such as standing, walking etc. Unlike the three cases above, here we use activity change information to obtain changepoint locations.
\end{itemize}

From this input dataset, we leave out 68 time series for evaluation, and use the rest of the data to construct a training dataset, comprising 563 labeled examples. Figure \ref{figure:tsne}(b) shows the distributions of the evaluation data and the generated training data. This is a stronger result that of the previous section, where both the input and generated data were created by anomaly injection. In this case, even with real time series with much more complex distributions, the distribution of the generated data contains the real data. The generated data also contains points in regions of the space where there exist no real data. This might imply that we are considering a larger range of anomalies than what the real data exhibit. 

Table \ref{table:real} shows the F-scores on the evaluation data. AutoML with hyperparameter tuning still performs the best. The low F-scores compared to the synthetic data indicate that anomaly detection on the real data is a considerably harder problem. StatSig comes out as the worst performing algorithm, which is expected because of its simplistic assumptions. BOCPD is seen to perform well in all the three cases which is not surprising, given that it came out as a top algorithm in a previous comprehensive study of changepoint algorithms \cite{van2020evaluation}. The fact that AutoML is able to consistently beat BOCPD in all our experiments implies that the use of the AutoML approach will be an important toolkit in changepoint detection problems. Another important point to emphasize is that, in these three experiments, comparing only among the base algorithms, three separate base algorithms performed the best. Among the base algorithms, Outlier performed the best for purely synthetic data, BOCPD performed the best for real data with anomaly injection, and MKDetector performed the best with real data. Thus, the best performing algorithm depends a lot on the charactistics of the input data. However, AutoML still manages to outperform each algorithm in all cases. Users have no way to figure out apriori which algorithm will work best for their dataset, and AutoML is extremely well suited for this scenario.

\begin{table}
\begin{center}
\begin{tabular}{|c|c|c|}
\toprule
Method & F-Score(Optimized) & F-Score(Random)\\
\midrule
AutoML & \textbf{0.498} & 0.415\\
Outlier & 0.287 & 0.3\\
CUSUM & 0.263 & 0.389\\
BOCPD & 0.352 & 0.186\\
Statsig & 0.184 & 0.288\\
MKDetector & 0.468 & 0.323\\
Prophet & 0.3 & 0.315\\
\bottomrule
\end{tabular}
\end{center}
\caption{Comparison of AutoML with baseline algorithms on Real Data}
\label{table:real}
\end{table}

\begin{table}
\begin{center}
\begin{tabular}{|c|c|c|c|}
\toprule
Method & Spike & Level & Trend\\
\midrule
Outlier & 25 & 3 & 11\\
\hline
CUSUM & 0 & 3 & 0\\
BOCPD & 0 & 17 & 0\\
Statsig & 0 & 2 & 2\\
\hline
MKDetector & 0 & 1 & 3\\
Prophet & 0 & 0 & 15\\
\bottomrule
\end{tabular}
\end{center}
\caption{Best models for each type of injected anomaly for evaluation data on synthetic data.}
\label{table:intuitive_synthetic}
\end{table}

\begin{table}
\begin{center}
\begin{tabular}{|c|c|c|c|}
\toprule
Method & Spike & Level & Trend\\
\midrule
Outlier & 22 & 7 & 1\\
\hline
CUSUM & 5 & 4 & 0\\
BOCPD & 1 & 12 & 0\\
Statsig & 0 & 1 & 5\\
\hline
MKDetector & 0 & 5 & 10\\
Prophet & 0 & 0 & 8\\
\bottomrule
\end{tabular}
\end{center}
\caption{Best models for each type of injected anomaly for evaluation data, on real data with injected anomalies}
\label{table:intuitive_injected}
\end{table}

\section{Conclusion}
In this paper, we describe an end-to-end system that we designed and implemented, to make smart algorithm and parameter selections for time series anomaly detection. This will help individual teams in large organizations set up time series monitoring scalably without expert guidance and achieve superior performance.  Our end-to-end system consists of multiple stages: 
\begin{itemize}
\item Building a labeled training dataset, which comes from a similar distribution as the unlabeled input data.
\item Using supervised learning to predict the best algorithm and parameters for a time series based on time series features
\end{itemize}

Using this AutoML approach, we are able to achieve high F-scores on synthetic and real datasets, consistently outperforming our base algorithms, which are the most standard well-known solutions used for anomaly and changepoint detection. The code for this system is open sourced and available publicly. 

In the future, we would like to explore more realistic methods for synthetic data generation based on generative models, such as GANs. Incorporating more intelligent search such as Bayesian Optimization methods will also lead to decrease in time to train the datasets. Another natural extension of our work is intent-based detection, where we construct training data based on natural language intents expressed by the users. For example, users may be interested only in changing trends and ignore spikes altogether.

\begin{acks}
We thank Rachel Pinkster, Karthik Kambatla and Aditya Priyadarshi for a number of discussions, 
which helped us improve our approach.
\end{acks}

\bibliographystyle{ACM-Reference-Format}
\bibliography{draft1}


\begin{thebibliography}{44}

%%% the Web address.

\ifx \showCODEN    \undefined \def \showCODEN     #1{\unskip}     \fi
\ifx \showDOI      \undefined \def \showDOI       #1{#1}\fi
\ifx \showISBNx    \undefined \def \showISBNx     #1{\unskip}     \fi
\ifx \showISBNxiii \undefined \def \showISBNxiii  #1{\unskip}     \fi
\ifx \showISSN     \undefined \def \showISSN      #1{\unskip}     \fi
\ifx \showLCCN     \undefined \def \showLCCN      #1{\unskip}     \fi
\ifx \shownote     \undefined \def \shownote      #1{#1}          \fi
\ifx \showarticletitle \undefined \def \showarticletitle #1{#1}   \fi
\ifx \showURL      \undefined \def \showURL       {\relax}        \fi
% invisible to TeX
\providecommand\bibfield[2]{#2}
\providecommand\bibinfo[2]{#2}
\providecommand\natexlab[1]{#1}
\providecommand\showeprint[2][]{arXiv:#2}

\bibitem[\protect\citeauthoryear{Adams and MacKay}{Adams and MacKay}{2007}]%
        {adams2007bayesian}
\bibfield{author}{\bibinfo{person}{Ryan~Prescott Adams} {and}
  \bibinfo{person}{David~JC MacKay}.} \bibinfo{year}{2007}\natexlab{}.
\newblock \showarticletitle{Bayesian online changepoint detection}.
\newblock \bibinfo{journal}{\emph{arXiv preprint arXiv:0710.3742}}
  (\bibinfo{year}{2007}).
\newblock


\bibitem[\protect\citeauthoryear{Aminikhanghahi and Cook}{Aminikhanghahi and
  Cook}{2017}]%
        {aminikhanghahi2017survey}
\bibfield{author}{\bibinfo{person}{Samaneh Aminikhanghahi} {and}
  \bibinfo{person}{Diane~J Cook}.} \bibinfo{year}{2017}\natexlab{}.
\newblock \showarticletitle{A survey of methods for time series change point
  detection}.
\newblock \bibinfo{journal}{\emph{Knowledge and information systems}}
  \bibinfo{volume}{51}, \bibinfo{number}{2} (\bibinfo{year}{2017}),
  \bibinfo{pages}{339--367}.
\newblock


\bibitem[\protect\citeauthoryear{Anguita, Ghio, Oneto, Parra, Reyes-Ortiz,
  et~al\mbox{.}}{Anguita et~al\mbox{.}}{2013}]%
        {anguita2013public}
\bibfield{author}{\bibinfo{person}{Davide Anguita}, \bibinfo{person}{Alessandro
  Ghio}, \bibinfo{person}{Luca Oneto}, \bibinfo{person}{Xavier Parra},
  \bibinfo{person}{Jorge~Luis Reyes-Ortiz}, {et~al\mbox{.}}}
  \bibinfo{year}{2013}\natexlab{}.
\newblock \showarticletitle{A public domain dataset for human activity
  recognition using smartphones.}. In \bibinfo{booktitle}{\emph{Esann}},
  Vol.~\bibinfo{volume}{3}. \bibinfo{pages}{3}.
\newblock


\bibitem[\protect\citeauthoryear{Bhatnagar, Kassianik, Liu, Lan, Yang, Cassius,
  Sahoo, Arpit, Subramanian, Woo, et~al\mbox{.}}{Bhatnagar
  et~al\mbox{.}}{2021}]%
        {bhatnagar2021merlion}
\bibfield{author}{\bibinfo{person}{Aadyot Bhatnagar}, \bibinfo{person}{Paul
  Kassianik}, \bibinfo{person}{Chenghao Liu}, \bibinfo{person}{Tian Lan},
  \bibinfo{person}{Wenzhuo Yang}, \bibinfo{person}{Rowan Cassius},
  \bibinfo{person}{Doyen Sahoo}, \bibinfo{person}{Devansh Arpit},
  \bibinfo{person}{Sri Subramanian}, \bibinfo{person}{Gerald Woo},
  {et~al\mbox{.}}} \bibinfo{year}{2021}\natexlab{}.
\newblock \showarticletitle{Merlion: A machine learning library for time
  series}.
\newblock \bibinfo{journal}{\emph{arXiv preprint arXiv:2109.09265}}
  (\bibinfo{year}{2021}).
\newblock


\bibitem[\protect\citeauthoryear{Braei and Wagner}{Braei and Wagner}{2020}]%
        {braei2020anomaly}
\bibfield{author}{\bibinfo{person}{Mohammad Braei} {and}
  \bibinfo{person}{Sebastian Wagner}.} \bibinfo{year}{2020}\natexlab{}.
\newblock \showarticletitle{Anomaly detection in univariate time-series: A
  survey on the state-of-the-art}.
\newblock \bibinfo{journal}{\emph{arXiv preprint arXiv:2004.00433}}
  (\bibinfo{year}{2020}).
\newblock


\bibitem[\protect\citeauthoryear{Chakraborty, Shah, Soltani, Swigart, Yang, and
  Buckingham}{Chakraborty et~al\mbox{.}}{2020}]%
        {chakraborty2020building}
\bibfield{author}{\bibinfo{person}{Sayan Chakraborty}, \bibinfo{person}{Smit
  Shah}, \bibinfo{person}{Kiumars Soltani}, \bibinfo{person}{Anna Swigart},
  \bibinfo{person}{Luyao Yang}, {and} \bibinfo{person}{Kyle Buckingham}.}
  \bibinfo{year}{2020}\natexlab{}.
\newblock \showarticletitle{Building an Automated and Self-Aware Anomaly
  Detection System}. In \bibinfo{booktitle}{\emph{2020 IEEE International
  Conference on Big Data (Big Data)}}. IEEE, \bibinfo{pages}{1465--1475}.
\newblock


\bibitem[\protect\citeauthoryear{Choudhary, Kejariwal, and Orsini}{Choudhary
  et~al\mbox{.}}{2017}]%
        {choudhary2017runtime}
\bibfield{author}{\bibinfo{person}{Dhruv Choudhary}, \bibinfo{person}{Arun
  Kejariwal}, {and} \bibinfo{person}{Francois Orsini}.}
  \bibinfo{year}{2017}\natexlab{}.
\newblock \showarticletitle{On the runtime-efficacy trade-off of anomaly
  detection techniques for real-time streaming data}.
\newblock \bibinfo{journal}{\emph{arXiv preprint arXiv:1710.04735}}
  (\bibinfo{year}{2017}).
\newblock


\bibitem[\protect\citeauthoryear{Cleland, Han, Nugent, Lee, McClean, Zhang, and
  Lee}{Cleland et~al\mbox{.}}{2014}]%
        {cleland2014evaluation}
\bibfield{author}{\bibinfo{person}{Ian Cleland}, \bibinfo{person}{Manhyung
  Han}, \bibinfo{person}{Chris Nugent}, \bibinfo{person}{Hosung Lee},
  \bibinfo{person}{Sally McClean}, \bibinfo{person}{Shuai Zhang}, {and}
  \bibinfo{person}{Sungyoung Lee}.} \bibinfo{year}{2014}\natexlab{}.
\newblock \showarticletitle{Evaluation of prompted annotation of activity data
  recorded from a smart phone}.
\newblock \bibinfo{journal}{\emph{Sensors}} \bibinfo{volume}{14},
  \bibinfo{number}{9} (\bibinfo{year}{2014}), \bibinfo{pages}{15861--15879}.
\newblock


\bibitem[\protect\citeauthoryear{Cleveland, Cleveland, McRae, and
  Terpenning}{Cleveland et~al\mbox{.}}{1990}]%
        {cleveland1990stl}
\bibfield{author}{\bibinfo{person}{Robert~B Cleveland},
  \bibinfo{person}{William~S Cleveland}, \bibinfo{person}{Jean~E McRae}, {and}
  \bibinfo{person}{Irma Terpenning}.} \bibinfo{year}{1990}\natexlab{}.
\newblock \showarticletitle{STL: A seasonal-trend decomposition}.
\newblock \bibinfo{journal}{\emph{J. Off. Stat}} \bibinfo{volume}{6},
  \bibinfo{number}{1} (\bibinfo{year}{1990}), \bibinfo{pages}{3--73}.
\newblock


\bibitem[\protect\citeauthoryear{Cook, M{\i}s{\i}rl{\i}, and Fan}{Cook
  et~al\mbox{.}}{2019}]%
        {cook2019anomaly}
\bibfield{author}{\bibinfo{person}{Andrew~A Cook}, \bibinfo{person}{G{\"o}ksel
  M{\i}s{\i}rl{\i}}, {and} \bibinfo{person}{Zhong Fan}.}
  \bibinfo{year}{2019}\natexlab{}.
\newblock \showarticletitle{Anomaly detection for IoT time-series data: A
  survey}.
\newblock \bibinfo{journal}{\emph{IEEE Internet of Things Journal}}
  \bibinfo{volume}{7}, \bibinfo{number}{7} (\bibinfo{year}{2019}),
  \bibinfo{pages}{6481--6494}.
\newblock


\bibitem[\protect\citeauthoryear{De~Vito, Massera, Piga, Martinotto, and
  Di~Francia}{De~Vito et~al\mbox{.}}{2008}]%
        {de2008field}
\bibfield{author}{\bibinfo{person}{Saverio De~Vito}, \bibinfo{person}{Ettore
  Massera}, \bibinfo{person}{Marco Piga}, \bibinfo{person}{Luca Martinotto},
  {and} \bibinfo{person}{Girolamo Di~Francia}.}
  \bibinfo{year}{2008}\natexlab{}.
\newblock \showarticletitle{On field calibration of an electronic nose for
  benzene estimation in an urban pollution monitoring scenario}.
\newblock \bibinfo{journal}{\emph{Sensors and Actuators B: Chemical}}
  \bibinfo{volume}{129}, \bibinfo{number}{2} (\bibinfo{year}{2008}),
  \bibinfo{pages}{750--757}.
\newblock


\bibitem[\protect\citeauthoryear{Dua and Graff}{Dua and Graff}{2017}]%
        {Dua:2019}
\bibfield{author}{\bibinfo{person}{Dheeru Dua} {and} \bibinfo{person}{Casey
  Graff}.} \bibinfo{year}{2017}\natexlab{}.
\newblock \bibinfo{title}{{UCI} Machine Learning Repository}.
\newblock
\newblock
\urldef\tempurl%
\url{http://archive.ics.uci.edu/ml}
\showURL{%
\tempurl}


\bibitem[\protect\citeauthoryear{Fearnhead and Rigaill}{Fearnhead and
  Rigaill}{2019}]%
        {fearnhead2019changepoint}
\bibfield{author}{\bibinfo{person}{Paul Fearnhead} {and}
  \bibinfo{person}{Guillem Rigaill}.} \bibinfo{year}{2019}\natexlab{}.
\newblock \showarticletitle{Changepoint detection in the presence of outliers}.
\newblock \bibinfo{journal}{\emph{J. Amer. Statist. Assoc.}}
  \bibinfo{volume}{114}, \bibinfo{number}{525} (\bibinfo{year}{2019}),
  \bibinfo{pages}{169--183}.
\newblock


\bibitem[\protect\citeauthoryear{Freeman and Beaver}{Freeman and
  Beaver}{2019}]%
        {freeman2019human}
\bibfield{author}{\bibinfo{person}{Cynthia Freeman} {and} \bibinfo{person}{Ian
  Beaver}.} \bibinfo{year}{2019}\natexlab{}.
\newblock \bibinfo{title}{Human-in-the-Loop Selection of Optimal Time Series
  Anomaly Detection Methods}.
\newblock
\newblock


\bibitem[\protect\citeauthoryear{Google}{Google}{2021a}]%
        {google2021mobility}
\bibfield{author}{\bibinfo{person}{Google}.} \bibinfo{year}{2021}\natexlab{a}.
\newblock \bibinfo{title}{Google Mobility Data}.
\newblock
  \bibinfo{howpublished}{\url{https://www.google.com/covid19/mobility/}}.
\newblock
\newblock
\shownote{Accessed: 2022-01-06.}


\bibitem[\protect\citeauthoryear{Google}{Google}{2021b}]%
        {kaggle2017webtraffic}
\bibfield{author}{\bibinfo{person}{Google}.} \bibinfo{year}{2021}\natexlab{b}.
\newblock \bibinfo{title}{Kaggle Web Traffic Forecasting}.
\newblock
  \bibinfo{howpublished}{\url{https://www.kaggle.com/c/web-traffic-time-series-forecasting}}.
\newblock
\newblock
\shownote{Accessed: 2022-01-06.}


\bibitem[\protect\citeauthoryear{Hyndman and Khandakar}{Hyndman and
  Khandakar}{2008}]%
        {hyndman2008automatic}
\bibfield{author}{\bibinfo{person}{Rob~J Hyndman} {and}
  \bibinfo{person}{Yeasmin Khandakar}.} \bibinfo{year}{2008}\natexlab{}.
\newblock \showarticletitle{Automatic time series forecasting: the forecast
  package for R}.
\newblock \bibinfo{journal}{\emph{Journal of statistical software}}
  \bibinfo{volume}{27}, \bibinfo{number}{1} (\bibinfo{year}{2008}),
  \bibinfo{pages}{1--22}.
\newblock


\bibitem[\protect\citeauthoryear{Iskandaryan, Ramos, and Trilles}{Iskandaryan
  et~al\mbox{.}}{2020}]%
        {iskandaryan2020air}
\bibfield{author}{\bibinfo{person}{Ditsuhi Iskandaryan},
  \bibinfo{person}{Francisco Ramos}, {and} \bibinfo{person}{Sergio Trilles}.}
  \bibinfo{year}{2020}\natexlab{}.
\newblock \showarticletitle{Air quality prediction in smart cities using
  machine learning technologies based on sensor data: A review}.
\newblock \bibinfo{journal}{\emph{Applied Sciences}} \bibinfo{volume}{10},
  \bibinfo{number}{7} (\bibinfo{year}{2020}), \bibinfo{pages}{2401}.
\newblock


\bibitem[\protect\citeauthoryear{Jamil, Kim, et~al\mbox{.}}{Jamil
  et~al\mbox{.}}{2021}]%
        {jamil2021ensemble}
\bibfield{author}{\bibinfo{person}{Faisal Jamil}, \bibinfo{person}{Dohyeun
  Kim}, {et~al\mbox{.}}} \bibinfo{year}{2021}\natexlab{}.
\newblock \showarticletitle{An Ensemble of a Prediction and Learning Mechanism
  for Improving Accuracy of Anomaly Detection in Network Intrusion
  Environments}.
\newblock \bibinfo{journal}{\emph{Sustainability}} \bibinfo{volume}{13},
  \bibinfo{number}{18} (\bibinfo{year}{2021}), \bibinfo{pages}{10057}.
\newblock


\bibitem[\protect\citeauthoryear{Kendall}{Kendall}{1948}]%
        {kendall1948rank}
\bibfield{author}{\bibinfo{person}{Maurice~George Kendall}.}
  \bibinfo{year}{1948}\natexlab{}.
\newblock \showarticletitle{Rank correlation methods.}
\newblock  (\bibinfo{year}{1948}).
\newblock


\bibitem[\protect\citeauthoryear{Killick, Fearnhead, and Eckley}{Killick
  et~al\mbox{.}}{2012}]%
        {killick2012optimal}
\bibfield{author}{\bibinfo{person}{Rebecca Killick}, \bibinfo{person}{Paul
  Fearnhead}, {and} \bibinfo{person}{Idris~A Eckley}.}
  \bibinfo{year}{2012}\natexlab{}.
\newblock \showarticletitle{Optimal detection of changepoints with a linear
  computational cost}.
\newblock \bibinfo{journal}{\emph{J. Amer. Statist. Assoc.}}
  \bibinfo{volume}{107}, \bibinfo{number}{500} (\bibinfo{year}{2012}),
  \bibinfo{pages}{1590--1598}.
\newblock


\bibitem[\protect\citeauthoryear{Knoblauch, Jewson, and Damoulas}{Knoblauch
  et~al\mbox{.}}{2018}]%
        {knoblauch2018doubly}
\bibfield{author}{\bibinfo{person}{Jeremias Knoblauch}, \bibinfo{person}{Jack~E
  Jewson}, {and} \bibinfo{person}{Theodoros Damoulas}.}
  \bibinfo{year}{2018}\natexlab{}.
\newblock \showarticletitle{Doubly Robust Bayesian Inference for Non-Stationary
  Streaming Data with beta-Divergences}. In \bibinfo{booktitle}{\emph{Advances
  in Neural Information Processing Systems}}. \bibinfo{pages}{64--75}.
\newblock


\bibitem[\protect\citeauthoryear{Kurt, Y{\i}ld{\i}z, Ceritli, Sankur, and
  Cemgil}{Kurt et~al\mbox{.}}{2018}]%
        {kurt2018bayesian}
\bibfield{author}{\bibinfo{person}{Bar{\i}{\c{s}} Kurt},
  \bibinfo{person}{{\c{C}}a{\u{g}}atay Y{\i}ld{\i}z},
  \bibinfo{person}{Taha~Yusuf Ceritli}, \bibinfo{person}{B{\"u}lent Sankur},
  {and} \bibinfo{person}{Ali~Taylan Cemgil}.} \bibinfo{year}{2018}\natexlab{}.
\newblock \showarticletitle{A Bayesian change point model for detecting
  SIP-based DDoS attacks}.
\newblock \bibinfo{journal}{\emph{Digital Signal Processing}}
  \bibinfo{volume}{77} (\bibinfo{year}{2018}), \bibinfo{pages}{48--62}.
\newblock


\bibitem[\protect\citeauthoryear{Lai, Zha, Wang, Xu, Zhao, Kumar, Chen,
  Zumkhawaka, Wan, Martinez, et~al\mbox{.}}{Lai et~al\mbox{.}}{2020}]%
        {lai2020tods}
\bibfield{author}{\bibinfo{person}{Kwei-Herng Lai}, \bibinfo{person}{Daochen
  Zha}, \bibinfo{person}{Guanchu Wang}, \bibinfo{person}{Junjie Xu},
  \bibinfo{person}{Yue Zhao}, \bibinfo{person}{Devesh Kumar},
  \bibinfo{person}{Yile Chen}, \bibinfo{person}{Purav Zumkhawaka},
  \bibinfo{person}{Minyang Wan}, \bibinfo{person}{Diego Martinez},
  {et~al\mbox{.}}} \bibinfo{year}{2020}\natexlab{}.
\newblock \showarticletitle{TODS: An Automated Time Series Outlier Detection
  System}.
\newblock \bibinfo{journal}{\emph{arXiv preprint arXiv:2009.09822}}
  (\bibinfo{year}{2020}).
\newblock


\bibitem[\protect\citeauthoryear{Lavielle and Teyssiere}{Lavielle and
  Teyssiere}{2007}]%
        {lavielle2007adaptive}
\bibfield{author}{\bibinfo{person}{Marc Lavielle} {and} \bibinfo{person}{Gilles
  Teyssiere}.} \bibinfo{year}{2007}\natexlab{}.
\newblock \showarticletitle{Adaptive detection of multiple change-points in
  asset price volatility}.
\newblock In \bibinfo{booktitle}{\emph{Long memory in economics}}.
  \bibinfo{publisher}{Springer}, \bibinfo{pages}{129--156}.
\newblock


\bibitem[\protect\citeauthoryear{Lavin and Ahmad}{Lavin and Ahmad}{2015}]%
        {lavin2015evaluating}
\bibfield{author}{\bibinfo{person}{Alexander Lavin} {and}
  \bibinfo{person}{Subutai Ahmad}.} \bibinfo{year}{2015}\natexlab{}.
\newblock \showarticletitle{Evaluating real-time anomaly detection
  algorithms--the Numenta anomaly benchmark}. In \bibinfo{booktitle}{\emph{2015
  IEEE 14th International Conference on Machine Learning and Applications
  (ICMLA)}}. IEEE, \bibinfo{pages}{38--44}.
\newblock


\bibitem[\protect\citeauthoryear{LeDell and Poirier}{LeDell and
  Poirier}{2020}]%
        {ledell2020h2o}
\bibfield{author}{\bibinfo{person}{Erin LeDell} {and}
  \bibinfo{person}{Sebastien Poirier}.} \bibinfo{year}{2020}\natexlab{}.
\newblock \showarticletitle{H2o automl: Scalable automatic machine learning}.
  In \bibinfo{booktitle}{\emph{Proceedings of the AutoML Workshop at ICML}},
  Vol.~\bibinfo{volume}{2020}.
\newblock


\bibitem[\protect\citeauthoryear{Li, Zha, Venugopal, Zou, and Hu}{Li
  et~al\mbox{.}}{2020}]%
        {li2020pyodds}
\bibfield{author}{\bibinfo{person}{Yuening Li}, \bibinfo{person}{Daochen Zha},
  \bibinfo{person}{Praveen Venugopal}, \bibinfo{person}{Na Zou}, {and}
  \bibinfo{person}{Xia Hu}.} \bibinfo{year}{2020}\natexlab{}.
\newblock \showarticletitle{Pyodds: An end-to-end outlier detection system with
  automated machine learning}. In \bibinfo{booktitle}{\emph{Companion
  Proceedings of the Web Conference 2020}}. \bibinfo{pages}{153--157}.
\newblock


\bibitem[\protect\citeauthoryear{Mann}{Mann}{1945}]%
        {mann1945nonparametric}
\bibfield{author}{\bibinfo{person}{Henry~B Mann}.}
  \bibinfo{year}{1945}\natexlab{}.
\newblock \showarticletitle{Nonparametric tests against trend}.
\newblock \bibinfo{journal}{\emph{Econometrica: Journal of the econometric
  society}} (\bibinfo{year}{1945}), \bibinfo{pages}{245--259}.
\newblock


\bibitem[\protect\citeauthoryear{Manogaran and Lopez}{Manogaran and
  Lopez}{2018}]%
        {manogaran2018spatial}
\bibfield{author}{\bibinfo{person}{Gunasekaran Manogaran} {and}
  \bibinfo{person}{Daphne Lopez}.} \bibinfo{year}{2018}\natexlab{}.
\newblock \showarticletitle{Spatial cumulative sum algorithm with big data
  analytics for climate change detection}.
\newblock \bibinfo{journal}{\emph{Computers \& Electrical Engineering}}
  \bibinfo{volume}{65} (\bibinfo{year}{2018}), \bibinfo{pages}{207--221}.
\newblock


\bibitem[\protect\citeauthoryear{Martin, Fowlkes, Tal, and Malik}{Martin
  et~al\mbox{.}}{2001}]%
        {martin2001database}
\bibfield{author}{\bibinfo{person}{David Martin}, \bibinfo{person}{Charless
  Fowlkes}, \bibinfo{person}{Doron Tal}, {and} \bibinfo{person}{Jitendra
  Malik}.} \bibinfo{year}{2001}\natexlab{}.
\newblock \showarticletitle{A database of human segmented natural images and
  its application to evaluating segmentation algorithms and measuring
  ecological statistics}. In \bibinfo{booktitle}{\emph{Proceedings Eighth IEEE
  International Conference on Computer Vision. ICCV 2001}},
  Vol.~\bibinfo{volume}{2}. IEEE, \bibinfo{pages}{416--423}.
\newblock


\bibitem[\protect\citeauthoryear{Meta}{Meta}{2021}]%
        {kats2021}
\bibfield{author}{\bibinfo{person}{Meta}.} \bibinfo{year}{2021}\natexlab{}.
\newblock \bibinfo{booktitle}{\emph{Kats}}.
\newblock
\urldef\tempurl%
\url{https://github.com/facebookresearch/Kats}
\showURL{%
\tempurl}


\bibitem[\protect\citeauthoryear{Oh, Rehg, Balch, and Dellaert}{Oh
  et~al\mbox{.}}{2008}]%
        {oh2008learning}
\bibfield{author}{\bibinfo{person}{Sang~Min Oh}, \bibinfo{person}{James~M
  Rehg}, \bibinfo{person}{Tucker Balch}, {and} \bibinfo{person}{Frank
  Dellaert}.} \bibinfo{year}{2008}\natexlab{}.
\newblock \showarticletitle{Learning and inferring motion patterns using
  parametric segmental switching linear dynamic systems}.
\newblock \bibinfo{journal}{\emph{International Journal of Computer Vision}}
  \bibinfo{volume}{77}, \bibinfo{number}{1} (\bibinfo{year}{2008}),
  \bibinfo{pages}{103--124}.
\newblock


\bibitem[\protect\citeauthoryear{Page}{Page}{1954}]%
        {page1954continuous}
\bibfield{author}{\bibinfo{person}{Ewan~S Page}.}
  \bibinfo{year}{1954}\natexlab{}.
\newblock \showarticletitle{Continuous inspection schemes}.
\newblock \bibinfo{journal}{\emph{Biometrika}} \bibinfo{volume}{41},
  \bibinfo{number}{1/2} (\bibinfo{year}{1954}), \bibinfo{pages}{100--115}.
\newblock


\bibitem[\protect\citeauthoryear{Panda and Nayak}{Panda and Nayak}{2016}]%
        {panda2016automatic}
\bibfield{author}{\bibinfo{person}{Soumya~Priyadarsini Panda} {and}
  \bibinfo{person}{Ajit~Kumar Nayak}.} \bibinfo{year}{2016}\natexlab{}.
\newblock \showarticletitle{Automatic speech segmentation in syllable centric
  speech recognition system}.
\newblock \bibinfo{journal}{\emph{International Journal of Speech Technology}}
  \bibinfo{volume}{19}, \bibinfo{number}{1} (\bibinfo{year}{2016}),
  \bibinfo{pages}{9--18}.
\newblock


\bibitem[\protect\citeauthoryear{Petluri and Al-Masri}{Petluri and
  Al-Masri}{2018}]%
        {petluri2018web}
\bibfield{author}{\bibinfo{person}{Navyasree Petluri} {and}
  \bibinfo{person}{Eyhab Al-Masri}.} \bibinfo{year}{2018}\natexlab{}.
\newblock \showarticletitle{Web traffic prediction of wikipedia pages}. In
  \bibinfo{booktitle}{\emph{2018 IEEE International Conference on Big Data (Big
  Data)}}. IEEE, \bibinfo{pages}{5427--5429}.
\newblock


\bibitem[\protect\citeauthoryear{Ren, Xu, Wang, Yi, Huang, Kou, Xing, Yang,
  Tong, and Zhang}{Ren et~al\mbox{.}}{2019}]%
        {ren2019time}
\bibfield{author}{\bibinfo{person}{Hansheng Ren}, \bibinfo{person}{Bixiong Xu},
  \bibinfo{person}{Yujing Wang}, \bibinfo{person}{Chao Yi},
  \bibinfo{person}{Congrui Huang}, \bibinfo{person}{Xiaoyu Kou},
  \bibinfo{person}{Tony Xing}, \bibinfo{person}{Mao Yang}, \bibinfo{person}{Jie
  Tong}, {and} \bibinfo{person}{Qi Zhang}.} \bibinfo{year}{2019}\natexlab{}.
\newblock \showarticletitle{Time-series anomaly detection service at
  microsoft}. In \bibinfo{booktitle}{\emph{Proceedings of the 25th ACM SIGKDD
  International Conference on Knowledge Discovery \& Data Mining}}.
  \bibinfo{pages}{3009--3017}.
\newblock


\bibitem[\protect\citeauthoryear{Shende, Feijoo-Lorenzo, and Bokde}{Shende
  et~al\mbox{.}}{2021}]%
        {shende2021cleants}
\bibfield{author}{\bibinfo{person}{Mayur~Kishor Shende},
  \bibinfo{person}{Andres~E Feijoo-Lorenzo}, {and}
  \bibinfo{person}{Neeraj~Dhanraj Bokde}.} \bibinfo{year}{2021}\natexlab{}.
\newblock \showarticletitle{cleanTS: Automated (AutoML) Tool to Clean
  Univariate Time Series at Microscales}.
\newblock \bibinfo{journal}{\emph{arXiv preprint arXiv:2110.11815}}
  (\bibinfo{year}{2021}).
\newblock


\bibitem[\protect\citeauthoryear{Taylor and Letham}{Taylor and Letham}{2018}]%
        {taylor2018forecasting}
\bibfield{author}{\bibinfo{person}{Sean~J Taylor} {and}
  \bibinfo{person}{Benjamin Letham}.} \bibinfo{year}{2018}\natexlab{}.
\newblock \showarticletitle{Forecasting at scale}.
\newblock \bibinfo{journal}{\emph{The American Statistician}}
  \bibinfo{volume}{72}, \bibinfo{number}{1} (\bibinfo{year}{2018}),
  \bibinfo{pages}{37--45}.
\newblock


\bibitem[\protect\citeauthoryear{Tran, Sasikumar, Hennessy, O’Loughlin, and
  Morgan}{Tran et~al\mbox{.}}{2020}]%
        {tran2020interpreting}
\bibfield{author}{\bibinfo{person}{Tu~Hao Tran}, \bibinfo{person}{Suraj
  Sasikumar}, \bibinfo{person}{Annemarie Hennessy}, \bibinfo{person}{Aiden
  O’Loughlin}, {and} \bibinfo{person}{Lucy Morgan}.}
  \bibinfo{year}{2020}\natexlab{}.
\newblock \showarticletitle{Interpreting the effect of social restrictions on
  cases of COVID-19 using mobility data}.
\newblock \bibinfo{journal}{\emph{Med J. Aust}}  \bibinfo{volume}{1}
  (\bibinfo{year}{2020}), \bibinfo{pages}{2020--09}.
\newblock


\bibitem[\protect\citeauthoryear{Valdez-Vivas, Gocmen, Korotkov, Fang, Goenka,
  and Chen}{Valdez-Vivas et~al\mbox{.}}{2018}]%
        {valdez2018real}
\bibfield{author}{\bibinfo{person}{Martin Valdez-Vivas}, \bibinfo{person}{Caner
  Gocmen}, \bibinfo{person}{Andrii Korotkov}, \bibinfo{person}{Ethan Fang},
  \bibinfo{person}{Kapil Goenka}, {and} \bibinfo{person}{Sherry Chen}.}
  \bibinfo{year}{2018}\natexlab{}.
\newblock \showarticletitle{A Real-time Framework for Detecting Efficiency
  Regressions in a Globally Distributed Codebase}. In
  \bibinfo{booktitle}{\emph{Proceedings of the 24th ACM SIGKDD International
  Conference on Knowledge Discovery \& Data Mining}}.
  \bibinfo{pages}{821--829}.
\newblock


\bibitem[\protect\citeauthoryear{van~den Burg and Williams}{van~den Burg and
  Williams}{2020}]%
        {van2020evaluation}
\bibfield{author}{\bibinfo{person}{Gerrit~JJ van~den Burg} {and}
  \bibinfo{person}{Christopher~KI Williams}.} \bibinfo{year}{2020}\natexlab{}.
\newblock \showarticletitle{An Evaluation of Change Point Detection
  Algorithms}.
\newblock \bibinfo{journal}{\emph{arXiv preprint arXiv:2003.06222}}
  (\bibinfo{year}{2020}).
\newblock


\bibitem[\protect\citeauthoryear{Yoon, Jarrett, and Van~der Schaar}{Yoon
  et~al\mbox{.}}{2019}]%
        {yoon2019time}
\bibfield{author}{\bibinfo{person}{Jinsung Yoon}, \bibinfo{person}{Daniel
  Jarrett}, {and} \bibinfo{person}{Mihaela Van~der Schaar}.}
  \bibinfo{year}{2019}\natexlab{}.
\newblock \showarticletitle{Time-series generative adversarial networks}.
\newblock  (\bibinfo{year}{2019}).
\newblock


\bibitem[\protect\citeauthoryear{Zhang, Jiang, Holt, Laptev, Komurlu, Gao, and
  Yu}{Zhang et~al\mbox{.}}{2021}]%
        {zhang2021self}
\bibfield{author}{\bibinfo{person}{Peiyi Zhang}, \bibinfo{person}{Xiaodong
  Jiang}, \bibinfo{person}{Ginger~M Holt}, \bibinfo{person}{Nikolay~Pavlovich
  Laptev}, \bibinfo{person}{Caner Komurlu}, \bibinfo{person}{Peng Gao}, {and}
  \bibinfo{person}{Yang Yu}.} \bibinfo{year}{2021}\natexlab{}.
\newblock \showarticletitle{Self-supervised learning for fast and scalable time
  series hyper-parameter tuning}.
\newblock \bibinfo{journal}{\emph{arXiv preprint arXiv:2102.05740}}
  (\bibinfo{year}{2021}).
\newblock


\end{thebibliography}

\appendix

\section{Reproducing our Experiments}
The code is available publicly in the Kats project\footnote{\url {https://github.com/facebookresearch/Kats/tree/main/kats/detectors/meta_learning}}.
For each of the data sources used, here is a more detailed description, in order to reproduce the results.

\begin{itemize}
\item \textbf{Synthetic Data}: We start with 18 original time series. For the first ten time series, we add trend, seasonality and Gaussian noise.
Trends are random draws from $N(10, 5)$, weekly seasonality magnitude from $N(5,3)$ and noise values from $N(2,2)$. For the 5 time series from the  ARIMA model, 
we use $ARIMA(p=2, d=\alpha, q=2)$, where $\alpha \sim Ber(0.5)$, with the coefficients for AR being $[0.1, 0.5]$ and MA being $[0.4, 0.1]$. Two time series are i.i.d. normal from $N(3,1)$ and one time series is i.i.d $t_5(3,1)$. 
In order to inject anomalies, the intervals between anomalies/changepoints are drawn from $G(1/100)$. Spike z-scores are drawn from $N(20,1)$ with signs drawn from $Ber(0.9)$. Level shift magnitudes are drawn from $N(2,1)$ and trend shift magnitudes are drawn from $N(8, 1)$.

\item \textbf{Real Data With Injected Anomalies}: We use 13 time series from the Google Mobility dataset. These correspond to time spent by users in each US State, in various places such as residences, workplaces etc. The 13 datasets correspond to AL(workplaces), AL(retail), CA(residential), CA(retail), VA(residential), VA(transit), FL(workplaces), IN(workplaces), AK(workplaces), IN(grocery), IN(workplaces), UT(workplaces), MD(workplaces). We removed the first 100 datapoints, which contained a number of level shifts due to lockdowns. We hand annotated the anomalies in the rest of the time series. 
We use 5 time series from the bee-dance dataset. Bee-dance data is 6 dimensional, and we look at a univariate time series of the 6th dimension. We get 5 time series, from periods without many anomalies. These correspond to subsets at $[1000:1200]$,$[1730: 1860]$, $[1940:2130]$, $[2300:2730]$ and $[3200:3500]$. Anomaly injection on top of this dataset has the same parameters as the synthetic data.

\item \textbf{Real Data}: Our data consists of 106 time series. Of these, 45 time series are from wikipedia web traffic dataset, 21 from the scanline dataset, 19 from the airquality dataset, and  
21 time series from the HAR dataset. Other than the HAR dataset, we hand labeled the 3 other datasets. Since the HAR datasets are used for time series classification, we were able to use the timestamps of class change as changepoints. For the web traffic dataset, we looked at a list of the most popular pages and looked at the time series of their web traffic. For the scanline dataset we took scanlines at 20 different locations, and for the air quality dataset, we took subsets of size 500 from the following columns: PT08.S1(CO), NMHC(GT), PT08.S2(NMHC), PT08.S3(NOx). 

\end{itemize}

\end{document}